\documentclass{article}
\usepackage{iftex}\ifPDFTeX\pdfoutput=1\fi

\PassOptionsToPackage{numbers,sort&compress}{natbib}
\usepackage[preprint]{neurips_2024}
\makeatletter\renewcommand{\@notice}{}\makeatother

\usepackage[utf8]{inputenc}
\usepackage[T1]{fontenc}
\usepackage{amsmath}
\usepackage{amssymb}
\usepackage{amsfonts}
\usepackage{graphicx}
\usepackage{booktabs}
\usepackage{array}
\usepackage{multirow}
\usepackage{tabularx}
\usepackage{microtype}
\usepackage{xcolor}
\usepackage{xspace}
\usepackage{url}
\usepackage{tikz}
\usetikzlibrary{arrows.meta, positioning, calc, fit, shapes.geometric,
                decorations.pathreplacing, patterns, backgrounds}
\usepackage{hyperref}
\usepackage[capitalize,nameinlink]{cleveref}

\definecolor{hueA}{HTML}{2E4A6B}
\definecolor{hueAm}{HTML}{5D7293}
\definecolor{hueAl}{HTML}{8DA1B8}
\definecolor{hueAll}{HTML}{C1CCD9}

\definecolor{hueB}{HTML}{3E7A7E}
\definecolor{hueBm}{HTML}{6A9DA1}
\definecolor{hueBl}{HTML}{A8C6C8}

\definecolor{hueC}{HTML}{A47A3D}
\definecolor{hueCl}{HTML}{C9A677}

\definecolor{hueD}{HTML}{6B4E75}
\definecolor{hueDl}{HTML}{9B83A2}

\definecolor{hueE}{HTML}{B76A5B}
\definecolor{hueEl}{HTML}{D29185}

\definecolor{hueF}{HTML}{4E6B52}
\definecolor{hueFl}{HTML}{83A089}

\definecolor{emph}{HTML}{8B3A2E}

\definecolor{inkdark}{HTML}{1F2937}
\definecolor{inkmed}{HTML}{4B5563}
\definecolor{inklight}{HTML}{94A3B8}
\definecolor{bgwarm}{HTML}{F7F4EE}

\colorlet{primary}{hueA}
\colorlet{accentA}{hueB}
\colorlet{accentB}{hueC}
\colorlet{accentC}{hueD}
\colorlet{accentD}{hueE}
\colorlet{accentE}{hueF}

\tikzset{
  >=Stealth,
  font=\sffamily,
  every node/.style={inner sep=0pt, outer sep=0pt},
  arw/.style={->, line width=1.3pt, color=inkmed!75},
  thinarw/.style={->, line width=0.7pt, color=inkmed!60},
  modarw/.style={->, line width=1.3pt, color=hueD!80, dashed},
  lossarw/.style={->, dashed, line width=1pt, color=emph!75},
  imgframe/.style={draw=inkmed!60, line width=0.4pt, inner sep=1pt},
  layerblock/.style={draw, rounded corners=2pt, thick, line width=0.5pt,
                     fill=hueD!12, draw=hueD!70,
                     minimum width=0.9cm, minimum height=0.5cm, inner sep=2pt,
                     font=\tiny\sffamily\color{hueD!35!black}, align=center},
  opblock/.style={draw, rounded corners=2pt, thick, line width=0.5pt,
                  fill=hueD!22, draw=hueD!70,
                  minimum width=0.9cm, minimum height=0.55cm, inner sep=2pt,
                  font=\tiny\sffamily\bfseries\color{hueD!30!black}, align=center},
  pill/.style={draw, rounded corners=4pt, thick, fill=hueC!15, draw=hueC!70,
               minimum width=1.2cm, minimum height=0.45cm, inner sep=2pt,
               font=\scriptsize\sffamily\color{hueC!40!black}},
  twoline-pill/.style={draw, rounded corners=3pt, thick, line width=0.5pt,
                       minimum width=1.25cm, minimum height=0.75cm, inner sep=2pt,
                       align=center, font=\tiny\sffamily},
  headerstrip/.style={rounded corners=3pt, line width=0.7pt},
  sumnode/.style={draw, circle, thick, fill=hueD!15, draw=hueD!75,
                  minimum size=0.7cm, inner sep=0pt,
                  font=\normalsize\bfseries\color{hueD!55!black}},
  frozenborder/.style={hueA!55, line width=1pt, dashed, rounded corners=4pt},
}

\newcommand{\qualok}{{\color{hueF!60!black}\checkmark}}
\newcommand{\qualbad}{{\color{emph}\ensuremath{\times}}}
\newcommand{\qimg}[3]{\setlength{\fboxsep}{0pt}\setlength{\fboxrule}{0.4pt}%
  \fcolorbox{inkmed!45}{white}{\includegraphics[width=#2,height=#3,keepaspectratio]{figures/qual/#1.jpg}}}
\newcommand{\gtpill}[1]{\tikz[baseline=-0.55ex]{\node[fill=hueB!13,draw=hueB!55,rounded corners=2pt,%
  inner sep=2.5pt,font=\footnotesize\color{inkdark}]{#1};}}
\newcommand{\gapmark}{\tikz[baseline=-0.5ex]{\node[fill=hueB!22,draw=hueB!65,rounded corners=2pt,%
  inner sep=1.6pt,font=\scriptsize\bfseries\color{hueB!48!black}]{\,$\cdots$\,};}}
\newcommand{\maskbox}[1]{{\scriptsize\color{inkmed}masked sentence:}%
  \par\vspace{1.5pt}\noindent\begin{tikzpicture}\node[fill=hueB!14,draw=hueB!62,rounded corners=2.5pt,%
  line width=0.6pt,inner sep=4pt,text width=\dimexpr\linewidth-9pt\relax,align=left,%
  font=\scriptsize\color{inkdark}]{#1};\end{tikzpicture}\par\vspace{4pt}}
\newcommand{\fillbox}[4]{\begin{tikzpicture}[baseline=(fb.center)]\node(fb)[fill=#1,draw=#2,%
  rounded corners=2.5pt,line width=0.5pt,inner sep=3.5pt,text width=3.3cm,align=left,%
  font=\scriptsize\color{inkdark}]{#3~#4};\end{tikzpicture}}
\newcommand{\qframe}[1]{\par\smallskip\noindent
  \begin{tikzpicture}\node[draw=inkmed!28,line width=0.5pt,rounded corners=3.5pt,fill=black!1.5,%
    inner sep=7pt]{\begin{minipage}{\dimexpr\linewidth-22pt\relax}#1\end{minipage}};\end{tikzpicture}%
  \par\smallskip}

\newif\ifshowtodo
\showtodotrue   

\newcommand{\dgemma}{\mbox{DiffusionGemma-26B}}
\newcommand{\argemma}{\mbox{Gemma-4-26B}}

\newcommand{\mimic}{\mbox{MIMIC-CXR}}

\newcommand{\eg}{e.g.\xspace}

\newcommand{\vs}{vs.\xspace}

\providecommand{\keywords}[1]{}
\let\subsubsection\paragraph

\title{Discrete Diffusion Language Models for Interactive Radiology Report Drafting}

\newcommand{\blfootnote}[1]{%
  \begingroup\renewcommand{\thefootnote}{}\footnote{#1}\endgroup%
  \addtocounter{footnote}{-1}}

\author{\normalfont
  \textbf{Max Van Puyvelde}\textsuperscript{*\,1,2} \qquad
  \textbf{H.~Ibrahim Gulluk}\textsuperscript{*\,3} \\
  \texttt{maxvpuyv@stanford.edu} \qquad \texttt{gulluk@stanford.edu} \\[5pt]
  \textbf{Wim Van Criekinge}\textsuperscript{\textdagger\,2} \qquad
  \textbf{Olivier Gevaert}\textsuperscript{\textdagger\,1} \\
  \texttt{wim.vancriekinge@ugent.be} \qquad \texttt{ogevaert@stanford.edu} \\[7pt]
  \textsuperscript{1}Department of Biomedical Data Science, Stanford University School of Medicine \\
  \textsuperscript{2}Department of Mathematical Modelling, Statistics \& Bioinformatics, Ghent University \\
  \textsuperscript{3}Department of Electrical Engineering, Stanford University
}

\begin{document}
\maketitle
\blfootnote{\textsuperscript{*}Joint first authors.\quad\textsuperscript{\textdagger}Joint senior authors.}

\begin{abstract}
Diffusion language models, which generate text by denoising a token canvas bidirectionally instead
of emitting tokens left to right, have become competitive with autoregressive (AR) generation.
Medical foundation models, however, remain almost entirely autoregressive. We adapt a
mixture-of-experts diffusion language model, \dgemma{}, and benchmark it against its same-size AR
sibling \argemma{} under an
identical LoRA recipe on medical visual question answering datasets, scored by a
verbosity-robust LLM judge. Diffusion matches or exceeds AR on all of them, and the finetuned model
($3.8$B active) is competitive with frontier vision-language models;
its decoding is also $3.5$--$4.4\times$ faster. Beyond this
parity, the diffusion model offers a drafting capability AR lacks: any-order infill. Because the
canvas is denoised bidirectionally, a radiologist can fix report fragments and have the model fill
the text between them, an operation inherent to diffusion but not to autoregression, which is
subpar at it.
This suits real reports, which are often terse or inconsistent across clinicians and institutions.

\end{abstract}

\section{Introduction}
\label{sec:intro}

Autoregressive (AR) generation, which produces text one token at a time from left to right,
underlies nearly all large language and vision-language models. Discrete diffusion language
models~\cite{d3pm2021,mdlm2024,llada2025} are a recent alternative: they generate a sequence by
iteratively denoising a fixed token canvas, with each position attending to the entire canvas rather
than only to preceding tokens. On general text these models are competitive with autoregressive
models of comparable size~\cite{llada2025,dream7b2025}, which makes them a plausible backbone for
domains that have so far relied on autoregression. One open instance,
\dgemma{}~\cite{diffusiongemma2026}, couples this denoising decoder with a native multimodal encoder,
and belongs to a model family that also includes a same-size autoregressive model,
\argemma{}~\cite{gemma4_2026}; the two share size, family, and lineage, and differ chiefly in their
generative paradigm.

Existing medical foundation models, however, are almost exclusively autoregressive. Radiology report
generation (RRG), the task of drafting a report from an image, is dominated by AR
models~\cite{maira1_2023,maira2_2024,rexrank2024,sdr2026,gulluk2026semenrich,openmedq2026}, as are medical vision-language
assistants~\cite{llavamed2023}. Whether a diffusion language model is viable as a medical foundation
model, both accurate enough and useful in the clinical workflow, is largely untested. A few diffusion models already generate
CXR reports~\cite{anchordiff2026,medim2025,echo2026}, but produce complete reports only and do not
address interactive drafting.

We finetune both the diffusion model and its autoregressive sibling on paired image-text data from
medical visual-question-answering datasets, under an identical LoRA recipe that varies only the
generative paradigm (same backbone size, vision tower, LoRA targets, and data), and benchmark them
against each other and frontier vision-language models with a verbosity-robust LLM judge.

Beyond accuracy, the two paradigms differ in what they can be conditioned on. Reporting practice
varies: negative and normal findings are stated explicitly in some settings and omitted in others,
and section conventions differ across institutions. A tool that completes or normalizes a report
around content the radiologist has already entered, at arbitrary positions, is therefore a useful
drafting operation. Because a diffusion decoder denoises the whole canvas bidirectionally, it can
fill such a gap from the fixed text on both sides. An autoregressive
model, conditioning each token only on preceding text, cannot: a fragment fixed after the gap cannot
inform the text filled before it. We call this any-order infill.

We make three contributions. (i)~\textbf{A diffusion language model is a competitive medical
foundation model.} \dgemma{} equals or exceeds its autoregressive sibling on medical VQA and
rivals frontier vision-language models while decoding $3.5$--$4.4\times$ faster
(\cref{sec:benchmark}, \cref{sec:speed}), in a matched comparison that varies only the generative
paradigm. (ii)~\textbf{Any-order infill
is a conditioning capability inherent to diffusion.} We cast infill as sampling a
report given fragments fixed at arbitrary positions (\cref{sec:method-infill}) and show on \mimic{}
that the diffusion model exploits context on both sides of a gap far more effectively than its
autoregressive sibling (\cref{sec:infill}). (iii)~We release our code and finetuned
checkpoints.\footnote{Code: \url{https://github.com/mxvp/discrete_diffusion_RRG}. Checkpoints:
\url{https://huggingface.co/gevaertlab/diffusiongemma-radiology-vqa}.}

\section{Related Work}
\label{sec:related}

\subsubsection{Diffusion for medical RRG and infill.}
RRG is dominated by autoregressive models such as MAIRA~\cite{maira1_2023,maira2_2024} and
ReXrank~\cite{rexrank2024}. Discrete
diffusion~\cite{d3pm2021,mdlm2024,llada2025} denoises a token canvas bidirectionally, and several
systems apply it to CXR report generation: \emph{AnchorDiff}~\cite{anchordiff2026}
(vision-conditioned LLaDA-8B, claimed as the first masked diffusion for RRG),
\emph{MeDiM}~\cite{medim2025} (unified any-to-any generation), and ECHO~\cite{echo2026} (one-step
distillation). All use bidirectionality only to improve full generation, and none isolate the
paradigm against a matched autoregressive backbone or expose interactive infill. Generic diffusion
infill is established~\cite{dream7b2025,dreamon2025} but not framed as clinical drafting, and
existing interactive report tools condition on a region~\cite{rgrg2023} or a
prefix~\cite{copilotcad2024}, not on fragments fixed at arbitrary positions.

\subsubsection{Medical VQA and LLM-as-judge.}
VQA-RAD~\cite{vqarad2018}, SLAKE~\cite{slake2021}, and VQA-Med~\cite{vqamed2019} pair radiology
images with short open- and closed-ended questions. Because exact-match scoring penalizes valid
paraphrases, open-ended medical VQA is now evaluated with an LLM judge~\cite{llavamed2023,mtbench2023},
which we adopt (\cref{sec:bench-metrics}).

\section{Method}
\label{sec:method}

\subsection{Matched Backbones}
\label{sec:method-backbones}

We compare diffusion and autoregression with everything else held fixed. The diffusion model is
\dgemma{}~\cite{diffusiongemma2026}, a discrete diffusion language model, and its AR sibling is
\argemma{}~\cite{gemma4_2026}; both are $25.2$B/$3.8$B-active mixture-of-experts models with a
SigLIP-lineage~\cite{siglip2023} vision encoder ($\sim$280 image tokens). We adapt each backbone with low-rank
adaptation (LoRA)~\cite{lora2022}: rank-$64$ updates ($\alpha{=}128$) on the attention and
shared-MLP projections, with the $128$ experts, the router, and the vision tower frozen. The experts
hold most of the weights, so adapting only the shared projections updates the model at a small
fraction of the cost of a full finetune, and the identical recipe and data across both backbones
leave the generative paradigm as the only deliberate variable. The optimizer is the lone exception: each paradigm keeps the
AdamW settings established for its objective,
since a shared one underfits one of the two losses. Full
hyperparameters are in \cref{sec:app-recipe}.

\subsection{Image-Conditioned Adaptation}
\label{sec:method-setup}

We condition on the image and diffuse the text target; the image is never generated. Both paradigms
are supervised only on the target tokens, with the image and prompt held fixed, and share the same
target string: the report (Findings and Impression) for drafting and infill, or a short answer for
VQA. A full report fits in one $256$-token canvas, so intra-report attention is bidirectional end to
end, which any-order infill requires.

Each paradigm is finetuned with its standard supervised objective, the only difference between the
two runs: the diffusion model uses the uniform-state dLLM
objective~\cite{d3pm2021,diffusiongemma2026} (a random fraction of the target tokens is replaced
with uniform draws from the vocabulary, a random token rather than a \texttt{[MASK]} symbol, and the
model is trained to recover them), and the AR sibling uses next-token cross-entropy on the same
targets.

\subsection{Any-Order Infill}
\label{sec:method-infill}

Infill fixes part of the report and has the model fill the rest, conditioned on what is fixed. A
radiologist who leaves a gap in a draft, for instance, fixes the text on either side of it. Let $F$
be the fixed positions, $\mathbf{a}$ the tokens placed there, and $\bar F$ the positions left to
fill; infill samples the free positions from the model's conditional
\begin{equation}
  \mathbf{x}_{\bar F}\,\sim\,p_\theta\!\big(\mathbf{x}_{\bar F}\,\mid\,\mathbf{x}_F{=}\mathbf{a},\,c\big),
  \label{eq:infill}
\end{equation}
its report distribution restricted to outputs that carry $\mathbf{a}$ at $F$. A diffusion decoder
samples this conditional directly, with no retraining: at each denoising step we re-impose $\mathbf{x}_F{=}\mathbf{a}$,
before the update so the model predicts the free positions while seeing the fixed ones, and after it
so they survive the step's re-randomization. Because attention within the canvas is bidirectional, a
free position conditions on fixed tokens to its right as much as to its left, so the gap is filled
from context on both sides.

An autoregressive model factors left to right, $p_\theta(\mathbf{x}\mid c)=\prod_i p_\theta(x_i\mid
x_{<i},c)$, and cannot sample this conditional: a token never sees the positions after it, so text
fixed after the gap cannot shape the fill before it. The paradigms differ in what they can be
conditioned on, not in writing quality; \cref{sec:infill} measures it, and \cref{sec:app-sampler}
gives the sampler.

\section{Experiments}
\label{sec:experiments}

We evaluate the adapted backbones three ways: medical VQA accuracy, inference speed, and any-order
infill.

\subsection{Medical VQA}
\label{sec:benchmark}
\label{sec:bench-metrics}

We compare diffusion and AR on medical VQA and place both against frontier vision-language models
(VLMs).
We evaluate on three medical-VQA datasets: VQA-RAD~\cite{vqarad2018}, SLAKE~\cite{slake2021}, and
VQA-Med-2019~\cite{vqamed2019} (sizes in \cref{sec:app-data}), each pairing an image and question
with a short open or closed answer.

We evaluate each backbone on every dataset both zero-shot (\emph{base}) and after per-dataset
finetuning (\emph{finetuned}), and compare against three frontier VLMs (Gemini-3.5-Flash, GPT-4.1-mini, and
Claude-Sonnet-4.6). Finetuning adapts the backbone with LoRA on the dataset (\cref{sec:app-recipe}).
The frontier models are run zero-shot in a single forward pass, without extended reasoning, and every model answers the same
$350$ held-out questions per dataset.

\begin{figure}[t]
  \centering
  \qframe{%
\begin{minipage}[c]{2.35cm}\centering\qimg{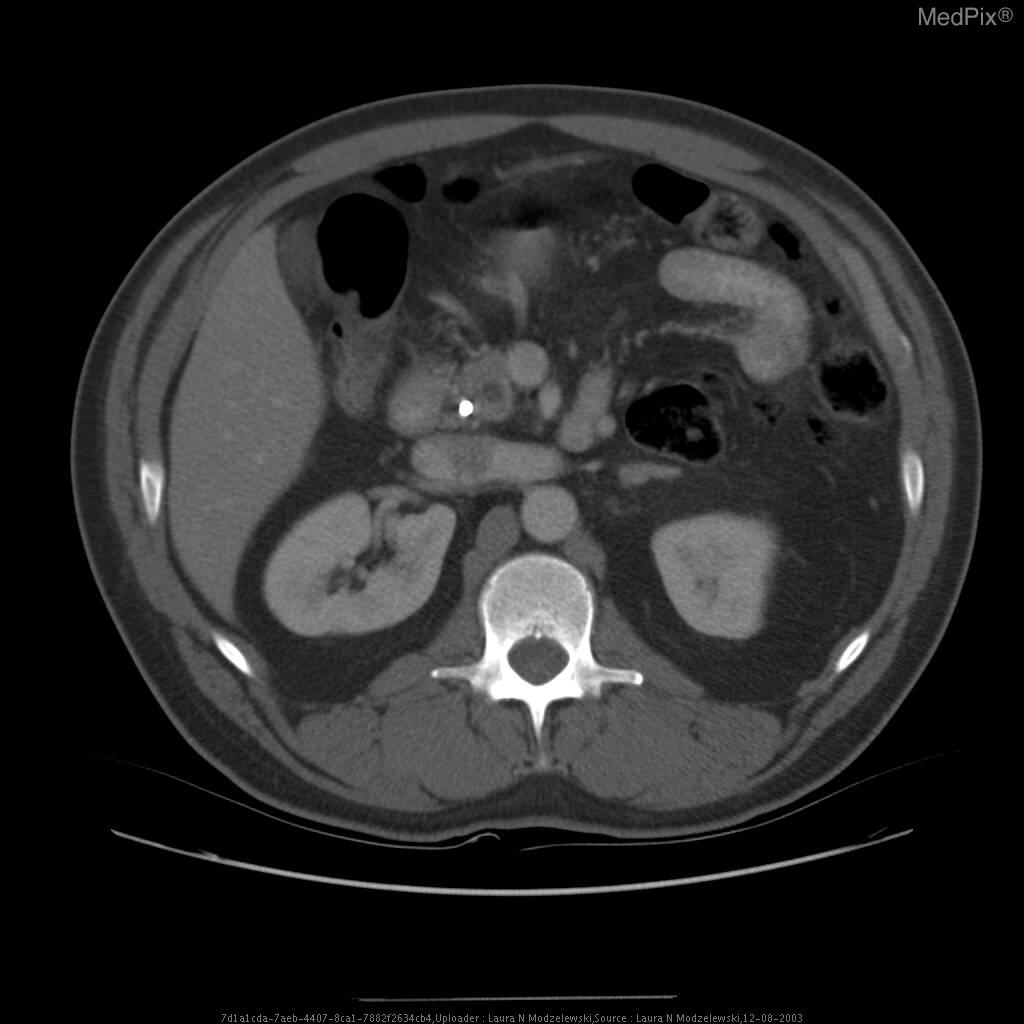}{2.2cm}{2.5cm}\end{minipage}\hfill
\begin{minipage}[c]{\dimexpr\linewidth-2.65cm\relax}
{\footnotesize\textbf{Q.}\ what is abnormal in the ct scan?}\par\vspace{4pt}
{\footnotesize\textbf{GT.}\ \gtpill{pancreatic ductal adenocarcinoma}}
\end{minipage}
\par\vspace{5pt}\hrulefill\par\vspace{4pt}
{\scriptsize\setlength{\tabcolsep}{4pt}\renewcommand{\arraystretch}{1.2}
\begin{tabularx}{\linewidth}{@{}p{1.75cm} c >{\raggedright\arraybackslash}X@{}}
    diff (base) & \qualbad & This axial CT scan of the abdomen shows a small, hyperdense (white) spot in the area of the common bile duct, which is consistent with a gallstone (choledocholithiasis).~[\ldots] \\
    AR (base) & \qualbad & This axial CT scan of the abdomen shows a high-density (bright white) object in the region of the pancreatic head/distal common bile duct, which is consistent with a gallstone or biliary stone.~[\ldots] \\
    diff (ft) & \qualok & pancreatic adenocarcinoma \\
    AR (ft) & \qualbad & This patient has a primary renal malignancy, with ultrasound and ct showing a large solid renal mass. On ultrasound, the mass is hypoechoic. On ct, the mass is hyperenhancing.~[\ldots] \\
    Gemini-3.5-Flash & \qualbad & Based on the CT scan, there is a calcified gallstone (a hyperdense, bright white calcification) located within the lumen of the gallbladder,~[\ldots] \\
    GPT-4.1-mini & \qualbad & The CT scan shows a metallic foreign body in the region of the pancreas, suggestive of a pancreatic stent. \\
    Sonnet-4.6 & \qualbad & The CT scan shows peripancreatic fat stranding and fluid surrounding the pancreas, consistent with acute pancreatitis.~[\ldots] \\
\end{tabularx}}
}

  \caption{\textbf{A medical-VQA example (VQA-Med).} Every model's answer to ``what is abnormal in
    the CT scan?'' (reference: \emph{pancreatic ductal adenocarcinoma}), with the LLM judge's verdict
    (\qualok~correct, \qualbad~incorrect). Base and frontier models reply in full sentences that
    exact-match scoring would reject regardless of content; here only the finetuned diffusion model
    answers correctly.}
  \label{fig:vqa-example}
\end{figure}

We score with an LLM judge. Standard exact-match accuracy is unsuitable for a cross-model
comparison here: base and frontier models answer in full sentences and score near zero regardless
of correctness (\cref{fig:vqa-example}). We therefore score semantic correctness: a fixed judge (Claude Sonnet 4.6) returns
a binary semantic-equivalence verdict per (question, reference, answer) triple, allowing paraphrase
and added explanation~\cite{llavamed2023,mtbench2023}, the standard for open-ended medical VQA.

\begin{figure}[tb]
  \centering
  \includegraphics[width=0.86\linewidth]{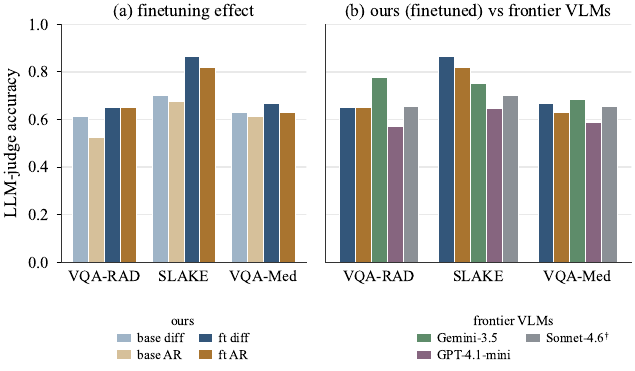}
  \caption{LLM-judge accuracy (Claude Sonnet 4.6). \textbf{(a)} base \vs
    finetuned, for diffusion and AR. \textbf{(b)} the finetuned $26$B model ($3.8$B active)
    against three frontier non-reasoning VLMs. $^{\dag}$Claude-Sonnet-4.6 is the judge model.}
  \label{fig:vqa-judge}
\end{figure}

\Cref{tab:vqa-main} reports LLM-judge accuracy for all models, and \cref{fig:vqa-judge} plots it.

\begin{table}[tb]
  \caption{LLM-judge accuracy (Claude Sonnet 4.6, semantic-equivalence), $n{=}350$ items per
    dataset. \emph{diff}\,/\,\emph{AR} are our two backbones (DiffusionGemma\,/\,Gemma-4), evaluated
    zero-shot (\emph{base}) and after per-dataset finetuning. Frontier VLMs (Gemini-3.5-Flash,
    GPT-4.1-mini, Claude-Sonnet-4.6) are zero-shot, single forward pass. Bold: best per dataset,
    \emph{separately} among our models and among the frontier VLMs. $^{\dag}$\,Claude-Sonnet-4.6 is
    also the judge model.}
  \label{tab:vqa-main}
  \centering
  \small
  \setlength{\tabcolsep}{5.5pt}
  \begin{tabular}{@{}l cc cc @{\hskip 0.8em}!{{\color{black!30}\vrule width 0.5pt}}@{\hskip 0.8em} ccc@{}}
    \toprule
    & \multicolumn{2}{c}{base} & \multicolumn{2}{c}{finetuned}
      & \multicolumn{3}{c}{frontier VLMs} \\
    \cmidrule(lr){2-3}\cmidrule(lr){4-5}\cmidrule(lr){6-8}
    Dataset & diff & AR & diff & AR & Gemini & GPT & Sonnet$^{\dag}$ \\
    \midrule
    VQA-RAD      & 0.614 & 0.523 & \textbf{0.649} & \textbf{0.649} & \textbf{0.777} & 0.571 & 0.654 \\
    SLAKE        & 0.700 & 0.674 & \textbf{0.863} & 0.817 & \textbf{0.751} & 0.646 & 0.703 \\
    VQA-Med-2019 & 0.629 & 0.614 & \textbf{0.666} & 0.631 & \textbf{0.683} & 0.586 & 0.654 \\
    \bottomrule
  \end{tabular}
\end{table}

\subsubsection{Finetuning.} Finetuning improves LLM-judge accuracy for both paradigms, most on
SLAKE: $+0.163$ diffusion ($0.700{\to}0.863$) and $+0.143$ AR ($0.674{\to}0.817$); VQA-RAD-AR
gains $+0.126$, and the VQA-Med gains are marginal. Base diffusion already reaches
$0.61$--$0.70$.

\subsubsection{Diffusion versus AR.} Finetuned diffusion equals or exceeds finetuned AR on the
judge metric for all three datasets, and base diffusion exceeds base AR on all three
(\cref{tab:vqa-main}). On per-item McNemar tests over the judge verdicts ($n{=}350$), the
difference is significant on SLAKE finetuned ($+0.046$, $p{=}0.026$) and VQA-RAD base ($+0.091$,
$p{<}0.001$); the other four diffusion-\vs-AR comparisons are not significant. The difference is concentrated on closed (yes/no) questions, where the
answer format is irrelevant (\eg on VQA-RAD finetuned, closed-question accuracy is $0.825$ for
diffusion \vs $0.757$ for AR). That a uniform-state denoising model matches its next-token sibling
at equal scale, on questions that turn on fine-grained image grounding, indicates the diffusion
paradigm is a viable substrate for a medical foundation model, on which the infill capability of
\cref{sec:infill} builds.

\subsubsection{Frontier VLMs.} The finetuned $26$B model ($3.8$B active) is competitive with the
three frontier VLMs (\cref{tab:vqa-main}; \cref{fig:vqa-judge}b). Finetuned diffusion has the
highest judge accuracy on SLAKE ($0.863$); Gemini-3.5-Flash is highest on VQA-RAD ($0.777$) and
VQA-Med ($0.683$). Finetuned diffusion exceeds GPT-4.1-mini on all three datasets; only
Gemini-3.5-Flash clearly surpasses it, on VQA-RAD and VQA-Med, while the judge model itself edges it
on VQA-RAD ($0.654$ \vs $0.649$, within noise at $n{=}350$). Example per-model answers appear in
\cref{sec:app-qual-vqa}.

\subsection{Inference Speed}
\label{sec:speed}

Latency matters for interactive drafting: the model must produce a draft fast enough to be
regenerated as the radiologist works. We characterize inference speed for the two decoders on
matched hardware.

The cost structures differ. AR cost scales with decoded tokens: each token is one sequential
forward pass (with KV caching), so latency grows with report length. Diffusion cost is set by the
denoising-step budget over the $256$-token canvas: each step is one forward pass updating all
unaccepted positions in parallel, independent of length. Because latency therefore depends on token
count and step budget rather than on the report's content, we measure a generic $\sim$$256$-token
generation rather than a specific dataset.

\dgemma{} drafts $3.5$--$4.4\times$ faster than its AR sibling and at
$5.7$--$7.1\times$ higher throughput (\cref{tab:speed}); AR is timed at its
natural, shorter output while diffusion fills the full canvas, so the comparison is generous to AR.

\begin{table}[tb]
  \caption{Inference speed on one H100 (bf16, $\sim$$256$-token generation). AR is greedy decode;
    DiffusionGemma is swept over the denoising-step budget. Speedup is latency relative to AR.}
  \label{tab:speed}
  \centering
  \small
  \setlength{\tabcolsep}{9pt}
  \begin{tabular}{@{}lccc@{}}
    \toprule
    Decoder & Latency (s) & Throughput (tok/s) & Speedup \\
    \midrule
    Gemma-4 (AR), greedy     & 6.43 & 24.6  & $1.0\times$ \\
    DiffusionGemma, 16 steps & 1.46 & 175.3 & $4.4\times$ \\
    DiffusionGemma, 32 steps & 1.74 & 147.4 & $3.7\times$ \\
    DiffusionGemma, 48 steps & 1.84 & 139.4 & $3.5\times$ \\
    \bottomrule
  \end{tabular}
\end{table}

\subsection{Any-Order Infill}
\label{sec:infill}

\begin{figure}[t]
  \centering
  \includegraphics[width=\linewidth]{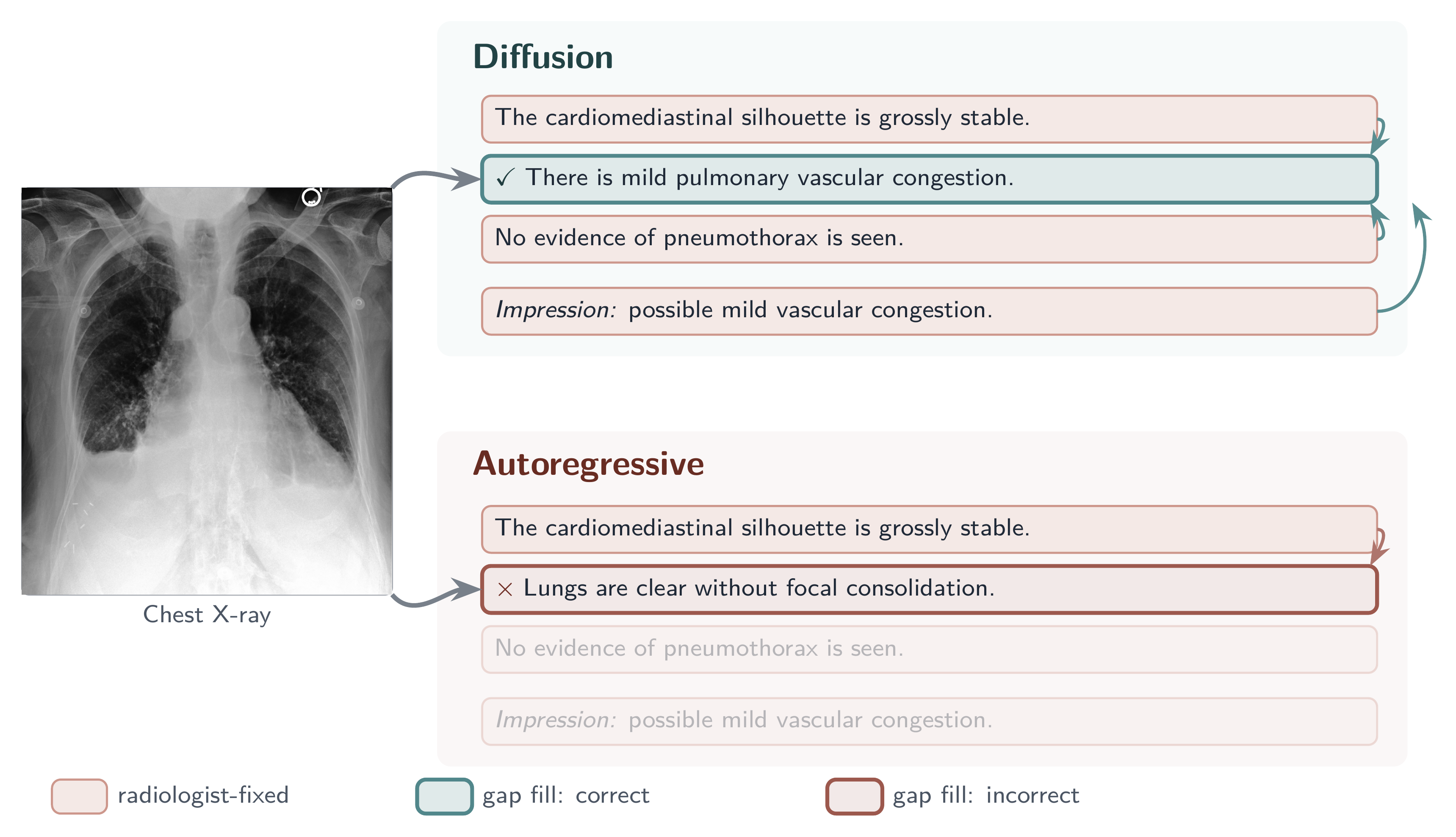}
  \caption{\textbf{Completing a gap from both sides.} One sentence of a chest X-ray report is masked
    (the gap) and filled from the surrounding fixed fragments. Top: the diffusion model draws on
    fragments on either side and recovers the sentence correctly. Bottom: the autoregressive sibling
    sees only the fragments before it, the rest greyed out, and reconstructs it incorrectly. Real
    \mimic{} example.}
  \label{fig:infill-concept}
\end{figure}

\Cref{sec:method-infill} cast infill as sampling the conditional of \cref{eq:infill}. Here we
evaluate the capability it affords that autoregression lacks: filling a gap in the report from the
fixed text on \emph{both} sides. A radiologist editing one sentence of a draft, for instance, wants
the surrounding text updated from both directions, where an AR model would regenerate only what
follows the edit. \Cref{fig:infill-concept} contrasts the two paradigms.

We mask one complete sentence (deterministically, near the middle) of each held-out
\mimic{}~\cite{mimiccxr2019} report ($n{=}249$, one canvas) and fill the resulting gap under two conditions, scoring each fill against the masked
sentence by token-F1 and by the LLM judge of \cref{sec:bench-metrics} (semantic equivalence to the
reference sentence). The \emph{bidirectional} condition clamps the fragments on both sides of the
gap; \emph{left-only} clamps only the left, emulating an AR view. We run this for the diffusion
model and for AR; in the AR \emph{bidirectional} condition the right-side context is supplied in the
prompt, the only way an autoregressive model can condition on it. The within-model gain from adding the
right context measures bidirectional exploitation, and the model$\times$context interaction
measures the capability asymmetry. Both models are the released (base) checkpoints, so the result
reflects architecture rather than finetuning.

\begin{table}[!t]
  \caption{Infill ablation on \mimic{} ($n{=}249$): fill a masked sentence with context on both
    sides \vs left only. Token-F1 and LLM-judge accuracy of the fill against the masked sentence;
    $\Delta$ is the gain from adding the right-side context (paired test). AR \emph{bidir.}\ is
    given both sides in its prompt. $^{\ast}p{<}10^{-3}$; n.s.: not significant.}
  \label{tab:infill}
  \centering
  \small
  \setlength{\tabcolsep}{7pt}
  \begin{tabular}{@{}l ccc @{\hskip 1em} ccc@{}}
    \toprule
    & \multicolumn{3}{c}{token-F1} & \multicolumn{3}{c}{LLM-judge} \\
    \cmidrule(lr){2-4}\cmidrule(lr){5-7}
    & bidir. & left & $\Delta$ & bidir. & left & $\Delta$ \\
    \midrule
    diffusion     & \textbf{0.320} & 0.211 & $+0.109^{\ast}$ & 0.285 & 0.157 & $+0.129^{\ast}$ \\
    AR (prompted) & 0.275 & 0.244 & $+0.031$\textsuperscript{\,n.s.} & 0.257 & 0.273 & $-0.016$\textsuperscript{\,n.s.} \\
    \bottomrule
  \end{tabular}
\end{table}

\Cref{tab:infill} reports the $2\times2$. The diffusion model uses the right-side context strongly:
adding it raises token-F1 by $+0.109$ (paired $t$, $p{<}10^{-10}$, $95\%$ CI $[+0.077,+0.141]$) and
judge accuracy by $+0.129$ ($p{=}2{\times}10^{-5}$). AR does not: even when prompted with both
sides, the right context does not significantly help it ($+0.031$ token-F1, $p{=}0.08$; $-0.016$
judge, n.s.). The model$\times$context interaction is significant on both metrics ($+0.078$
token-F1, $p{=}2{\times}10^{-4}$; $+0.145$ judge, $p{=}3{\times}10^{-4}$): diffusion benefits about
$3.5\times$ more from bidirectional context. Example fills are shown in
\cref{sec:app-qual-infill}.

\section{Conclusion}
\label{sec:conclusion}

We studied discrete diffusion versus autoregression for chest X-ray report drafting with two
same-size, same-family backbones, so the generative paradigm is the only variable. On a matched
medical-VQA benchmark scored by a verbosity-robust LLM judge, the diffusion model matches
or exceeds its AR sibling and is competitive with frontier vision-language models, while decoding
$3.5$--$4.4\times$ faster. Beyond this, it adds a drafting capability AR lacks, any-order infill: a
sampler modification lets a radiologist fix report fragments and have the diffusion model fill the
gaps between them. On \mimic{} it exploits context on both sides of a gap ($+0.109$ token-F1,
$+0.129$ judge accuracy) while AR does not, even when the right-side context is in its prompt.
We release our code and finetuned checkpoints.

\bibliographystyle{plainnat}
\bibliography{references}

\appendix
\renewcommand{\thetable}{A\arabic{table}}
\renewcommand{\thefigure}{A\arabic{figure}}
\renewcommand{\theHtable}{A\arabic{table}}
\renewcommand{\theHfigure}{A\arabic{figure}}
\setcounter{table}{0}
\setcounter{figure}{0}

\section{Backbones and Adaptation Recipe}
\label{sec:app-recipe}

\Cref{tab:models} lists the two backbones and their (identical) adaptation recipe; only the
generative paradigm and its established optimizer differ.

\begin{table}[h]
  \caption{The two backbones and their adaptation. Same family, size, vision tower, LoRA targets,
    and data (``same'' $=$ identical to the diffusion column); only the generative paradigm and
    its optimizer differ. Vision is frozen for both.}
  \label{tab:models}
  \centering
  \footnotesize
  \setlength{\tabcolsep}{6pt}
  \begin{tabular}{@{}lll@{}}
    \toprule
     & \dgemma{}-it & \argemma{}-it \\
    \midrule
    paradigm        & uniform-state discrete diffusion & autoregressive \\
    size            & 25.2B / 3.8B active MoE & same \\
    vision          & native \texttt{gemma4\_vision} ($\sim$$280$ tok) & same \\
    infill          & \textbf{yes} (bidirectional) & \textbf{no} (structural) \\
    \midrule
    LoRA            & $r64/\alpha128$, $\{q,k,v,o,\text{mlp}\}$ & same \\
    optimizer       & AdamW $\beta{=}(0.9,0.95)$ & AdamW $\beta{=}(0.9,0.999)$ \\
    learning rate   & $1\!\times\!10^{-4}$, cosine, warmup 100 & $1\!\times\!10^{-4}$, constant \\
    batch (grad-accum) & 16 & 24 \\
    diffusion settings & $\epsilon{=}0.1$, canvas $L{=}256$ & --- \\
    precision       & bf16, single H100 & same \\
    \bottomrule
  \end{tabular}
\end{table}

\section{Datasets}
\label{sec:app-data}

\Cref{tab:datasets} lists the three medical-VQA datasets and their sizes.

\begin{table}[h]
  \caption{Medical-VQA datasets. Sizes are train\,/\,test QA pairs; evaluation uses a fixed random
    subset of $n{=}350$ test items per dataset.}
  \label{tab:datasets}
  \centering
  \small
  \setlength{\tabcolsep}{9pt}
  \begin{tabular}{@{}lcl@{}}
    \toprule
    Dataset & Train\,/\,Test & Answers \\
    \midrule
    VQA-RAD~\cite{vqarad2018}      & 1{,}817 / 431      & yes/no, open ($\sim$1 word) \\
    SLAKE~\cite{slake2021}         & 4{,}919 / 1{,}053  & yes/no, open \\
    VQA-Med-2019~\cite{vqamed2019} & $\sim$12.8k / 2.0k & open: modality, plane, organ \\
    \bottomrule
  \end{tabular}
\end{table}

\section{Infill Sampler}
\label{sec:app-sampler}

We patch the uncompiled outer denoising step (the compiled inner sampler is reassigned as an
instance attribute and shadows a class-level patch), clamping the fixed positions on the incoming
canvas and on both outgoing canvases at each step (\cref{fig:infill}).

\begin{figure}[h]
  \centering
  \begin{minipage}{0.93\linewidth}\scriptsize
  \begin{verbatim}
  # clamp user-fixed tokens at fixed positions every denoising step
  # (patch the uncompiled outer _denoising_step, not the compiled inner sampler)
  def patched_step(self, *a, current_canvas=..., **kw):
      cc = where(fixed_mask, fixed_tokens, current_canvas)  # condition on fixed
      cur, argmax, *rest = orig_step(self, current_canvas=cc, ...)
      cur    = where(fixed_mask, fixed_tokens, cur)         # survive renoise
      argmax = where(fixed_mask, fixed_tokens, argmax)
      return (cur, argmax, *rest)
  \end{verbatim}
  \end{minipage}
  \caption{The any-order infill procedure (abridged). \texttt{fixed\_tokens} / \texttt{fixed\_mask}
    are $[B,L]$; the wrapper clamps the fixed positions on the incoming canvas and both outgoing
    canvases at each step.}
  \label{fig:infill}
\end{figure}

\section{Example Outputs}
\label{sec:app-qual}
\raggedbottom

\subsection{Medical VQA}
\label{sec:app-qual-vqa}
Held-out items from each dataset, with every model's answer and the LLM judge's verdict
(\qualok~correct, \qualbad~incorrect). Base and frontier models answer in full sentences, which
exact-match scoring penalizes regardless of correctness; the judge scores meaning
(\cref{sec:bench-metrics}). Long answers are abbreviated with [\ldots]; one sample per dataset is
shown in full to illustrate this verbosity.

\qframe{%
\begin{minipage}[c]{2.35cm}\centering\qimg{vqa_vqarad_1}{2.2cm}{2.5cm}\end{minipage}\hfill
\begin{minipage}[c]{\dimexpr\linewidth-2.65cm\relax}
{\footnotesize\textbf{Q.}\ Is there ventricular enlargement?}\par\vspace{4pt}
{\footnotesize\textbf{GT.}\ \gtpill{No}}
\end{minipage}
\par\vspace{5pt}\hrulefill\par\vspace{4pt}
{\scriptsize\setlength{\tabcolsep}{4pt}\renewcommand{\arraystretch}{1.2}
\begin{tabularx}{\linewidth}{@{}p{1.75cm} c >{\raggedright\arraybackslash}X@{}}
    diff (base) & \qualbad & Based on the provided MRI image, there is evidence of enlargement of the right lateral ventricle (the space on the left side of the image). The image shows a large area of high signal intensity (bright area) in the right hemisphere, which is consistent with a large acute ischemic stroke. The enlargement of the adjacent ventricle is likely due to "ex vacuo" effect, where the ventricle expands because of tissue loss (atrophy) from the stroke.~[\ldots] \\
    AR (base) & \qualbad & This image shows a Diffusion Weighted Imaging (DWI) MRI sequence of the brain. It is not possible to determine if there is ventricular enlargement from this single image. A formal evaluation of ventricular size requires several different types of MRI sequences (such as T1-weighted images) and a full examination of all slices in a study. The most striking finding on this image is a large, bright area in the right hemisphere (the left side of the image). In DWI, this bright signal is characteristic of an acute ischemic stroke affecting a large territory of the brain.~[\ldots] \\
    diff (ft) & \qualok & No \\
    AR (ft) & \qualbad & Yes \\
    Gemini-3.5-Flash & \qualbad & Based on the provided MRI scan, there is no generalized ventricular enlargement. However, there is asymmetric compression and effacement of the right lateral ventricle (shown on the left side of the image) due to mass effect from the large, hyperintense acute ischemic stroke in the right middle cerebral artery (MCA) territory. \\
    GPT-4.1-mini & \qualbad & Yes, there is ventricular enlargement. \\
    Sonnet-4.6 & \qualbad & Yes, there is ventricular enlargement (hydrocephalus) visible, with the right lateral ventricle appearing dilated. Additionally, there is a large area of restricted diffusion (bright signal on DWI) in the right hemisphere, consistent with an acute ischemic stroke involving the right MCA territory. \\
\end{tabularx}}
}

\qframe{%
\begin{minipage}[c]{2.35cm}\centering\qimg{vqa_vqarad_2}{2.2cm}{2.5cm}\end{minipage}\hfill
\begin{minipage}[c]{\dimexpr\linewidth-2.65cm\relax}
{\footnotesize\textbf{Q.}\ is there a mass}\par\vspace{4pt}
{\footnotesize\textbf{GT.}\ \gtpill{yes}}
\end{minipage}
\par\vspace{5pt}\hrulefill\par\vspace{4pt}
{\scriptsize\setlength{\tabcolsep}{4pt}\renewcommand{\arraystretch}{1.2}
\begin{tabularx}{\linewidth}{@{}p{1.75cm} c >{\raggedright\arraybackslash}X@{}}
    diff (base) & \qualok & This chest X-ray shows a rounded opacity (a white area) in the upper left lung.~[\ldots] \\
    AR (base) & \qualok & This chest X-ray shows a well-defined opacity (a "spot" or density) in the upper part of the left lung.~[\ldots] \\
    diff (ft) & \qualok & yes \\
    AR (ft) & \qualok & Yes \\
    Gemini-3.5-Flash & \qualbad & Based on the provided chest X-ray, there is no evidence of a definitive pulmonary mass. \\
    GPT-4.1-mini & \qualbad & No, there is no visible mass in this chest X-ray. \\
    Sonnet-4.6 & \qualbad & No, there is no discrete mass visible on this chest X-ray.~[\ldots] \\
\end{tabularx}}
}

\qframe{%
\begin{minipage}[c]{2.35cm}\centering\qimg{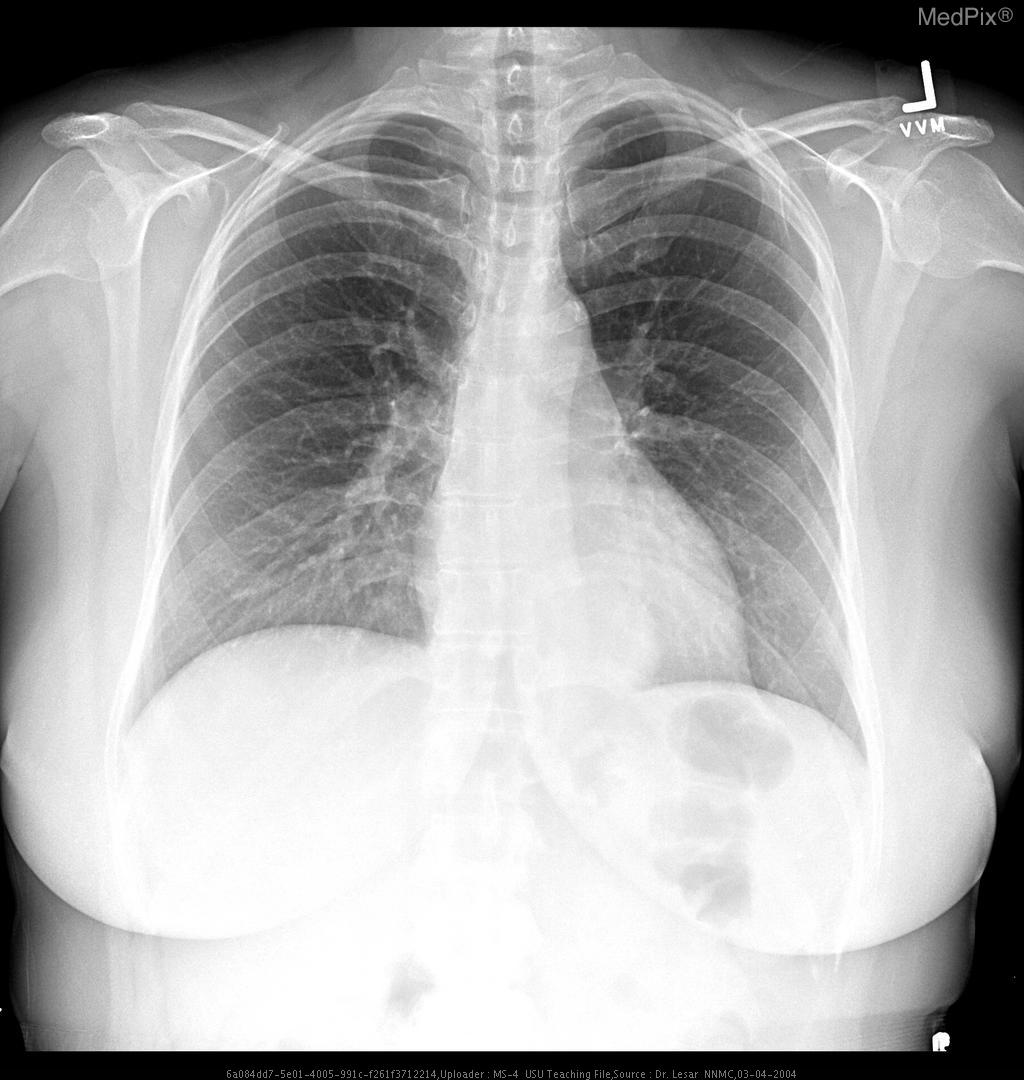}{2.2cm}{2.5cm}\end{minipage}\hfill
\begin{minipage}[c]{\dimexpr\linewidth-2.65cm\relax}
{\footnotesize\textbf{Q.}\ Describe the borders of the lesion.}\par\vspace{4pt}
{\footnotesize\textbf{GT.}\ \gtpill{Smooth}}
\end{minipage}
\par\vspace{5pt}\hrulefill\par\vspace{4pt}
{\scriptsize\setlength{\tabcolsep}{4pt}\renewcommand{\arraystretch}{1.2}
\begin{tabularx}{\linewidth}{@{}p{1.75cm} c >{\raggedright\arraybackslash}X@{}}
    diff (base) & \qualbad & The provided chest X-ray does not show any clearly defined lesions. The lung fields appear clear, and the heart size and shape are within normal limits.~[\ldots] \\
    AR (base) & \qualbad & I cannot provide a description of a lesion because no lesion is visible in this chest X-ray. The image appears to show normal anatomy for a chest radiograph.~[\ldots] \\
    diff (ft) & \qualok & Well-defined \\
    AR (ft) & \qualok & The lesion has well-defined borders. \\
    Gemini-3.5-Flash & \qualbad & Based on the provided chest radiograph, there is no visible lung lesion or abnormal mass.~[\ldots] \\
    GPT-4.1-mini & \qualbad & The borders of the lesion are irregular and poorly defined. \\
    Sonnet-4.6 & \qualbad & The lesion (located in the right hilum/perihilar region) has well-defined, lobulated borders, suggesting a discrete mass rather than an infiltrative process.~[\ldots] \\
\end{tabularx}}
}

\qframe{%
\begin{minipage}[c]{2.35cm}\centering\qimg{vqa_vqarad_4}{2.2cm}{2.5cm}\end{minipage}\hfill
\begin{minipage}[c]{\dimexpr\linewidth-2.65cm\relax}
{\footnotesize\textbf{Q.}\ From what structure is the large hypodense mass emanating from?}\par\vspace{4pt}
{\footnotesize\textbf{GT.}\ \gtpill{The base of the cecum}}
\end{minipage}
\par\vspace{5pt}\hrulefill\par\vspace{4pt}
{\scriptsize\setlength{\tabcolsep}{4pt}\renewcommand{\arraystretch}{1.2}
\begin{tabularx}{\linewidth}{@{}p{1.75cm} c >{\raggedright\arraybackslash}X@{}}
    diff (base) & \qualbad & Based on the provided CT scan of the abdomen, the large hypodense mass appears to be emanating from the small bowel.~[\ldots] \\
    AR (base) & \qualbad & This single axial CT image shows a large, hypodense (darker) mass in the central abdomen.~[\ldots] \\
    diff (ft) & \qualok & cecum \\
    AR (ft) & \qualbad & The mass is seen as an extension of the ascending colon. \\
    Gemini-3.5-Flash & \qualbad & Based on the axial CT image, the large hypodense mass in the lower abdomen/pelvis is emanating from the appendix (consistent with an appendi~[\ldots] \\
    GPT-4.1-mini & \qualbad & The large hypodense mass is emanating from the abdominal aorta. \\
    Sonnet-4.6 & \qualbad & The large hypodense mass is emanating from the mesentery (mesenteric origin). \\
\end{tabularx}}
}

\qframe{%
\begin{minipage}[c]{2.35cm}\centering\qimg{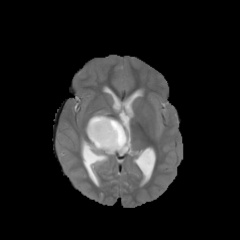}{2.2cm}{2.5cm}\end{minipage}\hfill
\begin{minipage}[c]{\dimexpr\linewidth-2.65cm\relax}
{\footnotesize\textbf{Q.}\ How many kinds of abnormalities are there in this image?}\par\vspace{4pt}
{\footnotesize\textbf{GT.}\ \gtpill{3}}
\end{minipage}
\par\vspace{5pt}\hrulefill\par\vspace{4pt}
{\scriptsize\setlength{\tabcolsep}{4pt}\renewcommand{\arraystretch}{1.2}
\begin{tabularx}{\linewidth}{@{}p{1.75cm} c >{\raggedright\arraybackslash}X@{}}
    diff (base) & \qualbad & Based on the single image provided, there is one primary area of abnormality visible: a large, hyperintense (bright) lesion located in the right hemisphere, specifically involving the thalamus/basal ganglia region. A definitive medical diagnosis requires a review of the entire imaging study, multiple sequences, and the patient's clinical history by a qualified professional. If this image belongs to you or someone you know, you should consult with a physician for a formal interpretation and clinical plan. \\
    AR (base) & \qualbad & Based on the provided image, which appears to be an axial MRI scan of a brain, there is a large, prominent abnormality visible. The image shows a significant area of high signal intensity (bright white area) located in the central/right region of the brain. This represents a single major type of abnormality, which could indicate several different medical conditions such as a tumor, hemorrhage, or an area of edema/infarction. Important Medical Disclaimer: I am an AI, not a doctor.~[\ldots] \\
    diff (ft) & \qualok & 3 \\
    AR (ft) & \qualok & There are 3 kinds of abnormalities in this image. \\
    Gemini-3.5-Flash & \qualbad & Based on the provided brain MRI, there are two main types of abnormalities visible: 1. A large intracranial mass/tumor (the well-defined, hyperintense lesion in the deep right hemisphere/thalamic region). 2. Obstructive ventriculomegaly/hydrocephalus (abnormal enlargement and dilation of the lateral ventricles, particularly the occipital and frontal horns, caused by mass effect and CSF obstruction). \\
    GPT-4.1-mini & \qualbad & There are two kinds of abnormalities in this image. \\
    Sonnet-4.6 & \qualbad & Looking at this brain MRI (DWI sequence), I can identify 2 kinds of abnormalities: 1. Restricted diffusion (bright signal) --- indicating acute ischemia/infarction in the posterior fossa/cerebellum region 2. Mass effect/enlarged structure --- suggesting associated edema or structural enlargement in that area \\
\end{tabularx}}
}

\qframe{%
\begin{minipage}[c]{2.35cm}\centering\qimg{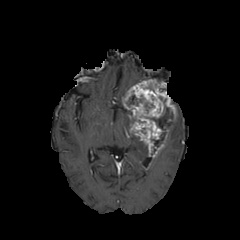}{2.2cm}{2.5cm}\end{minipage}\hfill
\begin{minipage}[c]{\dimexpr\linewidth-2.65cm\relax}
{\footnotesize\textbf{Q.}\ How many kinds of abnormalities are there in this image?}\par\vspace{4pt}
{\footnotesize\textbf{GT.}\ \gtpill{3}}
\end{minipage}
\par\vspace{5pt}\hrulefill\par\vspace{4pt}
{\scriptsize\setlength{\tabcolsep}{4pt}\renewcommand{\arraystretch}{1.2}
\begin{tabularx}{\linewidth}{@{}p{1.75cm} c >{\raggedright\arraybackslash}X@{}}
    diff (base) & \qualbad & Based on the image provided, there is one large, primary abnormality: a significant, heterogeneous mass located in the left hemisphere of the brain.~[\ldots] \\
    AR (base) & \qualbad & This image is a single medical scan (an axial slice of a brain MRI).~[\ldots] \\
    diff (ft) & \qualok & 3 \\
    AR (ft) & \qualbad & There is 1 kind of abnormality in this image. \\
    Gemini-3.5-Flash & \qualbad & Based on the provided brain MRI, there is 1 major kind of abnormality (a large, heterogeneous, enhancing intra-axial mass lesion in the left~[\ldots] \\
    GPT-4.1-mini & \qualbad & There is one kind of abnormality visible in this image. \\
    Sonnet-4.6 & \qualbad & Looking at this brain MRI image, I can identify 2 kinds of abnormalities: 1. Hemorrhage (hyperdense/bright area indicating blood) 2.~[\ldots] \\
\end{tabularx}}
}

\qframe{%
\begin{minipage}[c]{2.35cm}\centering\qimg{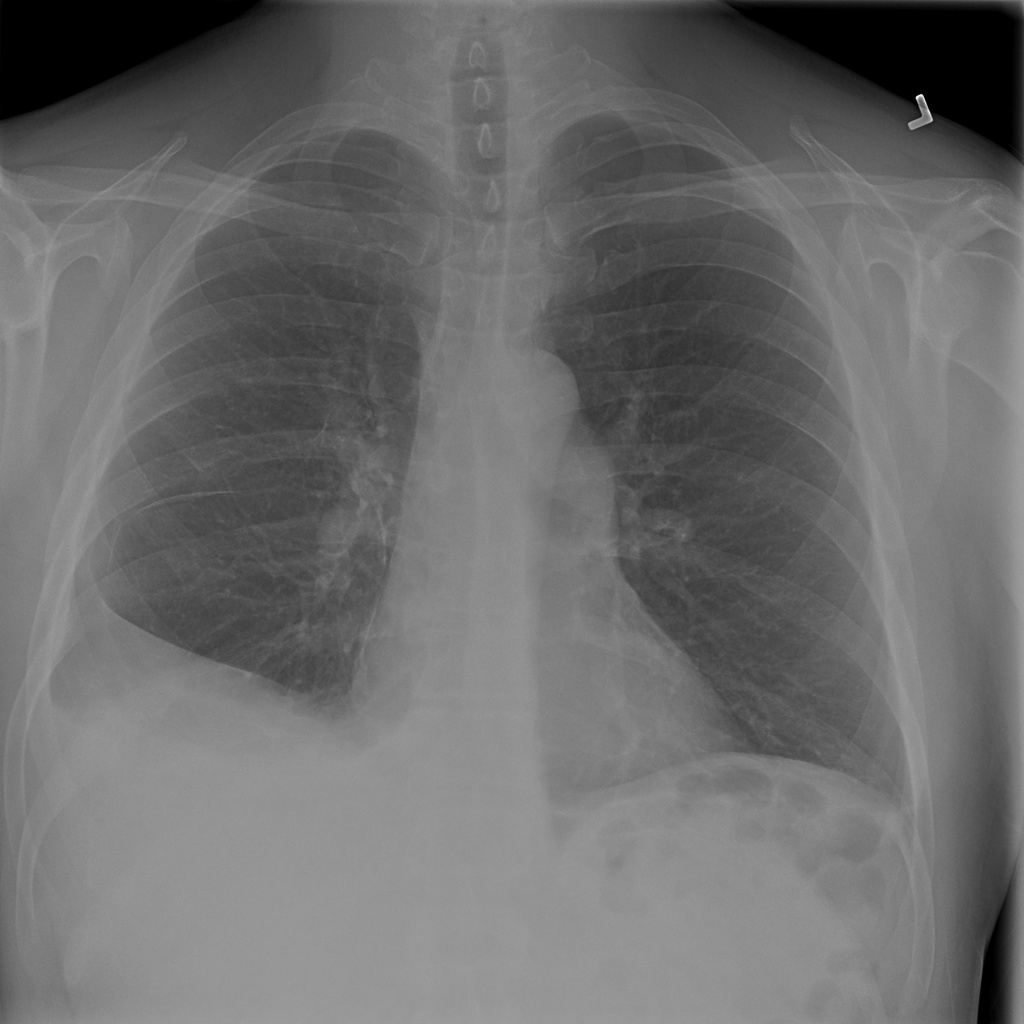}{2.2cm}{2.5cm}\end{minipage}\hfill
\begin{minipage}[c]{\dimexpr\linewidth-2.65cm\relax}
{\footnotesize\textbf{Q.}\ How many organs are there in this image?}\par\vspace{4pt}
{\footnotesize\textbf{GT.}\ \gtpill{2}}
\end{minipage}
\par\vspace{5pt}\hrulefill\par\vspace{4pt}
{\scriptsize\setlength{\tabcolsep}{4pt}\renewcommand{\arraystretch}{1.2}
\begin{tabularx}{\linewidth}{@{}p{1.75cm} c >{\raggedright\arraybackslash}X@{}}
    diff (base) & \qualbad & I cannot count the number of organs in this image. A chest X-ray is a medical imaging tool used to visualize internal structures, but it is not used to count organs.~[\ldots] \\
    AR (base) & \qualbad & I cannot provide a medical interpretation or diagnosis from this image.~[\ldots] \\
    diff (ft) & \qualok & 2 \\
    AR (ft) & \qualok & 2 \\
    Gemini-3.5-Flash & \qualbad & Based on this chest X-ray, we can identify 3 major visible organs: the lungs (left and right), the heart, and the stomach (visible via the g~[\ldots] \\
    GPT-4.1-mini & \qualbad & There are two organs visible in this image: the lungs. \\
    Sonnet-4.6 & \qualbad & In this chest X-ray, there are multiple organs visible, but the main identifiable ones are: 1. Lungs (2 --- left and right) 2. Heart 3. Liver (partially visible at the right lower border) 4.~[\ldots] \\
\end{tabularx}}
}

\qframe{%
\begin{minipage}[c]{2.35cm}\centering\qimg{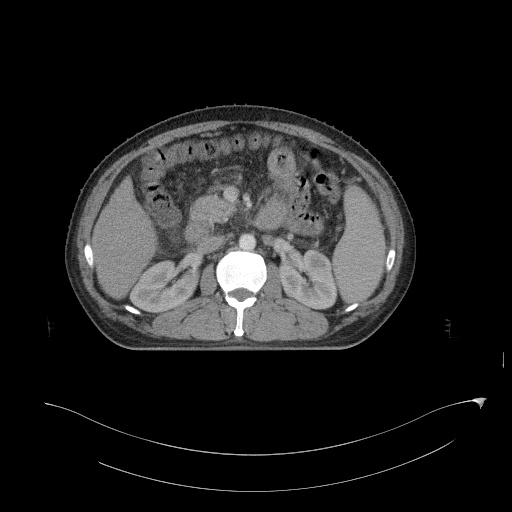}{2.2cm}{2.5cm}\end{minipage}\hfill
\begin{minipage}[c]{\dimexpr\linewidth-2.65cm\relax}
{\footnotesize\textbf{Q.}\ How many organs are there in this image?}\par\vspace{4pt}
{\footnotesize\textbf{GT.}\ \gtpill{6}}
\end{minipage}
\par\vspace{5pt}\hrulefill\par\vspace{4pt}
{\scriptsize\setlength{\tabcolsep}{4pt}\renewcommand{\arraystretch}{1.2}
\begin{tabularx}{\linewidth}{@{}p{1.75cm} c >{\raggedright\arraybackslash}X@{}}
    diff (base) & \qualbad & I cannot count the number of organs from a single CT scan slice. A CT scan consists of many sequential images that are necessary to fully identify and evaluate all internal structures.~[\ldots] \\
    AR (base) & \qualbad & I cannot provide a count of organs in this image. Interpreting medical imaging requires specialized training and context from a complete scan, which a single image cannot provide.~[\ldots] \\
    diff (ft) & \qualok & 6 \\
    AR (ft) & \qualok & There are 6 organs in this image. \\
    Gemini-3.5-Flash & \qualbad & Based on this abdominal CT slice, there are 4 major distinct organs visualized: 1. Spleen (on the right side of the image / anatomical left) 2.~[\ldots] \\
    GPT-4.1-mini & \qualbad & There are four organs visible in this image: the liver, both kidneys, and portions of the intestines. \\
    Sonnet-4.6 & \qualbad & In this abdominal CT scan, the visible organs include: 1. Kidneys (bilateral) --- 2 2. Bowel loops (small/large intestine) 3. Vertebral body (bone structure) 4.~[\ldots] \\
\end{tabularx}}
}

\qframe{%
\begin{minipage}[c]{2.35cm}\centering\qimg{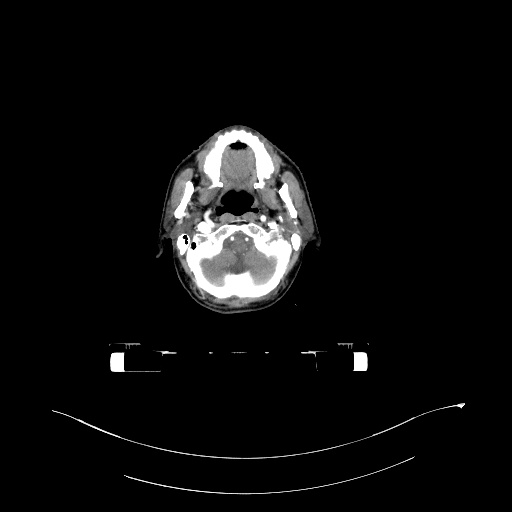}{2.2cm}{2.5cm}\end{minipage}\hfill
\begin{minipage}[c]{\dimexpr\linewidth-2.65cm\relax}
{\footnotesize\textbf{Q.}\ Is this a study of the neck?}\par\vspace{4pt}
{\footnotesize\textbf{GT.}\ \gtpill{No}}
\end{minipage}
\par\vspace{5pt}\hrulefill\par\vspace{4pt}
{\scriptsize\setlength{\tabcolsep}{4pt}\renewcommand{\arraystretch}{1.2}
\begin{tabularx}{\linewidth}{@{}p{1.75cm} c >{\raggedright\arraybackslash}X@{}}
    diff (base) & \qualbad & Yes, this is an axial CT scan of the lower head and upper neck region.~[\ldots] \\
    AR (base) & \qualbad & This single image is an axial CT slice through the level of the upper neck/base of the skull, showing structures such as the mandible, pharynx, and the base of the brain.~[\ldots] \\
    diff (ft) & \qualok & No \\
    AR (ft) & \qualok & No \\
    Gemini-3.5-Flash & \qualbad & Yes. \\
    GPT-4.1-mini & \qualbad & Yes. \\
    Sonnet-4.6 & \qualbad & Yes, this is a CT study of the neck/head junction region --- specifically an axial CT image at the level of the skull base/upper neck, showi~[\ldots] \\
\end{tabularx}}
}

\qframe{%
\begin{minipage}[c]{2.35cm}\centering\qimg{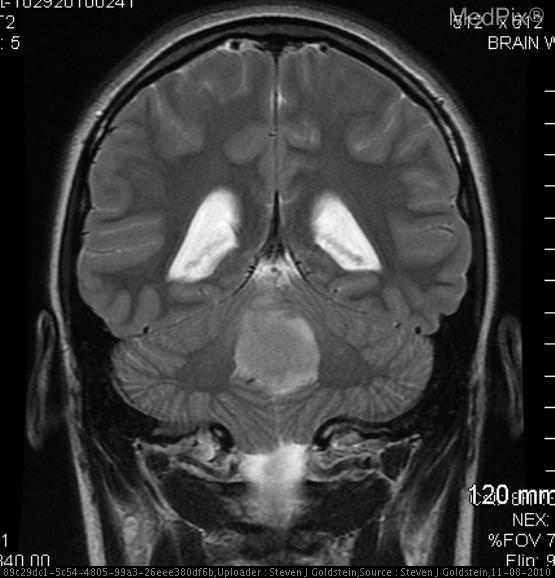}{2.2cm}{2.5cm}\end{minipage}\hfill
\begin{minipage}[c]{\dimexpr\linewidth-2.65cm\relax}
{\footnotesize\textbf{Q.}\ what is the primary abnormality in this image?}\par\vspace{4pt}
{\footnotesize\textbf{GT.}\ \gtpill{juvenile pilocytic astrocytoma}}
\end{minipage}
\par\vspace{5pt}\hrulefill\par\vspace{4pt}
{\scriptsize\setlength{\tabcolsep}{4pt}\renewcommand{\arraystretch}{1.2}
\begin{tabularx}{\linewidth}{@{}p{1.75cm} c >{\raggedright\arraybackslash}X@{}}
    diff (base) & \qualok & This T2-weighted coronal MRI of the brain shows a large, well-defined hyperintense (bright) area located in the cerebellum. The appearance and location of this lesion are concerning for a cerebellar tumor.~[\ldots] \\
    AR (base) & \qualbad & This image is a coronal T2-weighted MRI of the brain. The primary abnormality is a large, well-circumscribed, hyperintense (bright) lesion located in the midline of the cerebellum (the posterior fossa).~[\ldots] \\
    diff (ft) & \qualok & pillary astrocytoma \\
    AR (ft) & \qualok & The primary abnormality in this image is a juvenile pilocytic astrocytoma (jpa). Juvenile pilocytic astrocytoma is a benign, slow-growing neoplasm of the central nervous system.~[\ldots] \\
    Gemini-3.5-Flash & \qualbad & Based on the provided coronal T2-weighted brain MRI, the primary abnormality is a large, hyperintense mass in the brainstem (specifically wi~[\ldots] \\
    GPT-4.1-mini & \qualbad & The primary abnormality in this image is bilateral symmetrical hyperintense lesions in the basal ganglia, indicative of a metabolic or toxic~[\ldots] \\
    Sonnet-4.6 & \qualbad & The primary abnormality in this image is bilateral enlargement of the lateral ventricles (hydrocephalus).~[\ldots] \\
\end{tabularx}}
}

\qframe{%
\begin{minipage}[c]{2.35cm}\centering\qimg{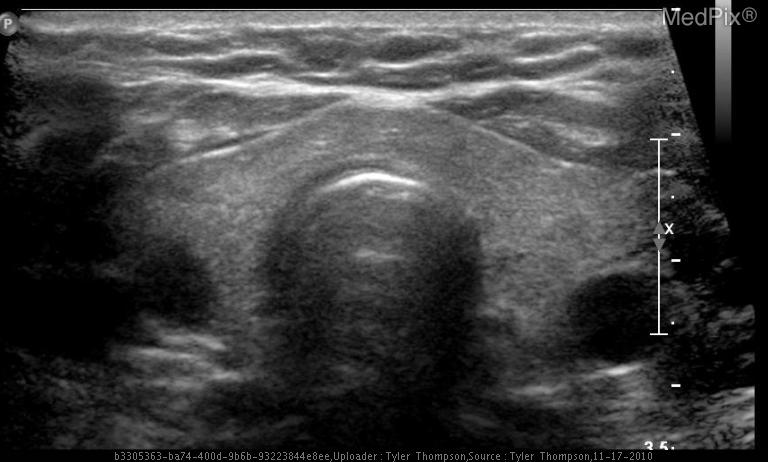}{2.2cm}{2.5cm}\end{minipage}\hfill
\begin{minipage}[c]{\dimexpr\linewidth-2.65cm\relax}
{\footnotesize\textbf{Q.}\ what is most alarming about this ultrasound?}\par\vspace{4pt}
{\footnotesize\textbf{GT.}\ \gtpill{papillary carcinoma of the thyroid}}
\end{minipage}
\par\vspace{5pt}\hrulefill\par\vspace{4pt}
{\scriptsize\setlength{\tabcolsep}{4pt}\renewcommand{\arraystretch}{1.2}
\begin{tabularx}{\linewidth}{@{}p{1.75cm} c >{\raggedright\arraybackslash}X@{}}
    diff (base) & \qualbad & This image shows an ultrasound of the thyroid gland. The most alarming feature in this image is the presence of a large, well-defined nodule in the right thyroid lobe (the darker/hypoechoic circular area on the left side of the image). While many thyroid nodules are benign, certain features on ultrasound can be associated with an increased risk of malignancy (cancer). These features, which radiologists look for, include:  Hypoechogenicity: The nodule appearing darker than the surrounding thyroid tissue.  Irregular margins: The edges of the nodule not being smooth or sharp.~[\ldots] \\
    AR (base) & \qualbad & An ultrasound image cannot be used to provide a diagnosis. A complete medical evaluation requires a review of the original high-resolution imaging, a physical examination, and a discussion of your medical history and symptoms. The image you provided is a single, low-resolution frame from a thyroid ultrasound. Interpreting such an image is limited because the most important features (such as the exact texture, borders, and vascularity of a potential nodule) require real-time scanning and specific measurement tools.~[\ldots] \\
    diff (ft) & \qualok & thyroid carcinoma \\
    AR (ft) & \qualbad & The ultrasound shows a normal thyroid gland, which is a reassuring finding in a patient with stable hypothyroidism. \\
    Gemini-3.5-Flash & \qualbad & Based on the provided thyroid/neck ultrasound, the most alarming finding is the absence of the right thyroid lobe (thyroid hemiagenesis). The image shows a normal-appearing left thyroid lobe and isthmus crossing anterior to the trachea, but the right thyroid bed (on the left side of the image) is empty, with no identifiable right thyroid lobe tissue lateral to the trachea. \\
    GPT-4.1-mini & \qualbad & The most alarming feature in this ultrasound is the presence of a large, well-defined, possibly cystic mass in the thyroid region, which could indicate a thyroid nodule or tumor requiring further evaluation. \\
    Sonnet-4.6 & \qualbad & The most alarming finding in this ultrasound is the irregular, hypoechoic solid mass with posterior acoustic shadowing, which raises concern for malignancy (such as breast carcinoma). The mass appears to have irregular margins and heterogeneous echotexture, features that are suspicious for a malignant lesion rather than a benign cyst or fibroadenoma. \\
\end{tabularx}}
}

\subsection{Any-Order Infill}
\label{sec:app-qual-infill}
Four held-out \mimic{} reports, each with one sentence masked. Its position is marked
(\gapmark{}) and the masked sentence is shown in the teal box below; the four fills form
a $\{$diffusion, AR$\}\times\{$bidirectional, left-only$\}$ grid. \emph{Bidirectional} supplies the
fixed text on both sides of the gap (for AR, in its prompt), \emph{left-only} only the left. Only the
bidirectional diffusion fill reconstructs the masked sentence; the others, including AR with both
sides in its prompt, cannot condition on the right-side context.

\qframe{%
\begin{minipage}[t]{2.35cm}\vspace{0pt}\centering\qimg{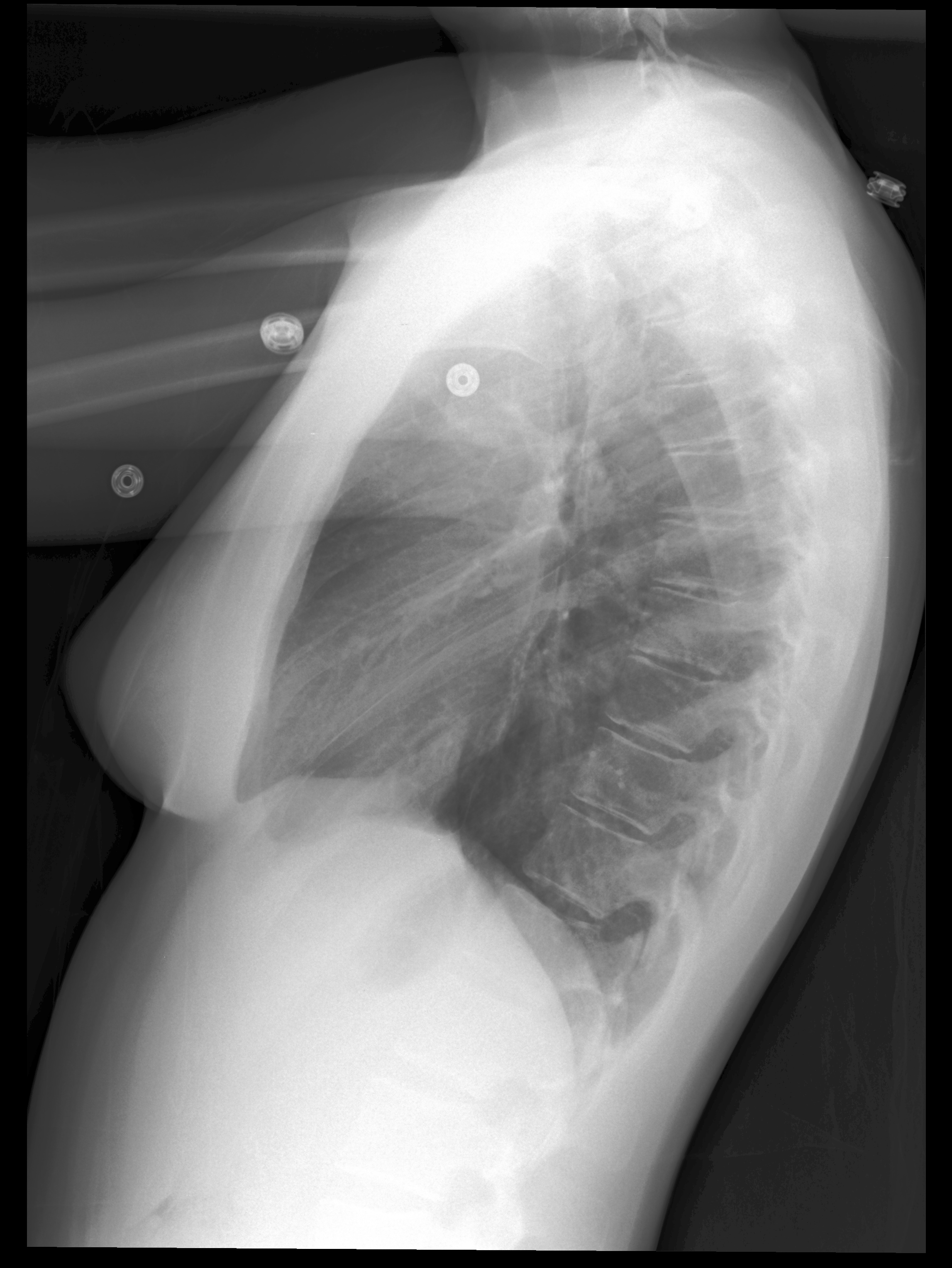}{2.25cm}{2.8cm}\end{minipage}\hfill
\begin{minipage}[t]{\dimexpr\linewidth-2.65cm\relax}\vspace{0pt}
{\scriptsize\textbf{Report.}\ The heart is normal in size. The mediastinal and hilar contours appear within normal limits. \gapmark{} The lungs appear clear. Bony structures are unremarkable. No evidence of acute disease.}
\par\vspace{5pt}\maskbox{There is no pleural effusion or pneumothorax.}
{\scriptsize\setlength{\tabcolsep}{4pt}\renewcommand{\arraystretch}{1.2}
\begin{tabular}{@{}r c c@{}}
& bidirectional & left-only \\[2pt]
\textbf{diffusion} & \fillbox{hueB!12}{hueB!60}{\qualok}{There is no pleural effusion or pneumothorax.} & \fillbox{hueE!8}{hueE!45}{\qualbad}{The lungs are clear, with no infiltrates,} \\[3pt]
\textbf{AR} & \fillbox{hueE!8}{hueE!45}{\qualbad}{The pleural spaces are clear.} & \fillbox{hueE!8}{hueE!45}{\qualbad}{The trachea is midline and the retrosternal clear space is preserved.} \\
\end{tabular}}
\end{minipage}
}

\qframe{%
\begin{minipage}[t]{2.35cm}\vspace{0pt}\centering\qimg{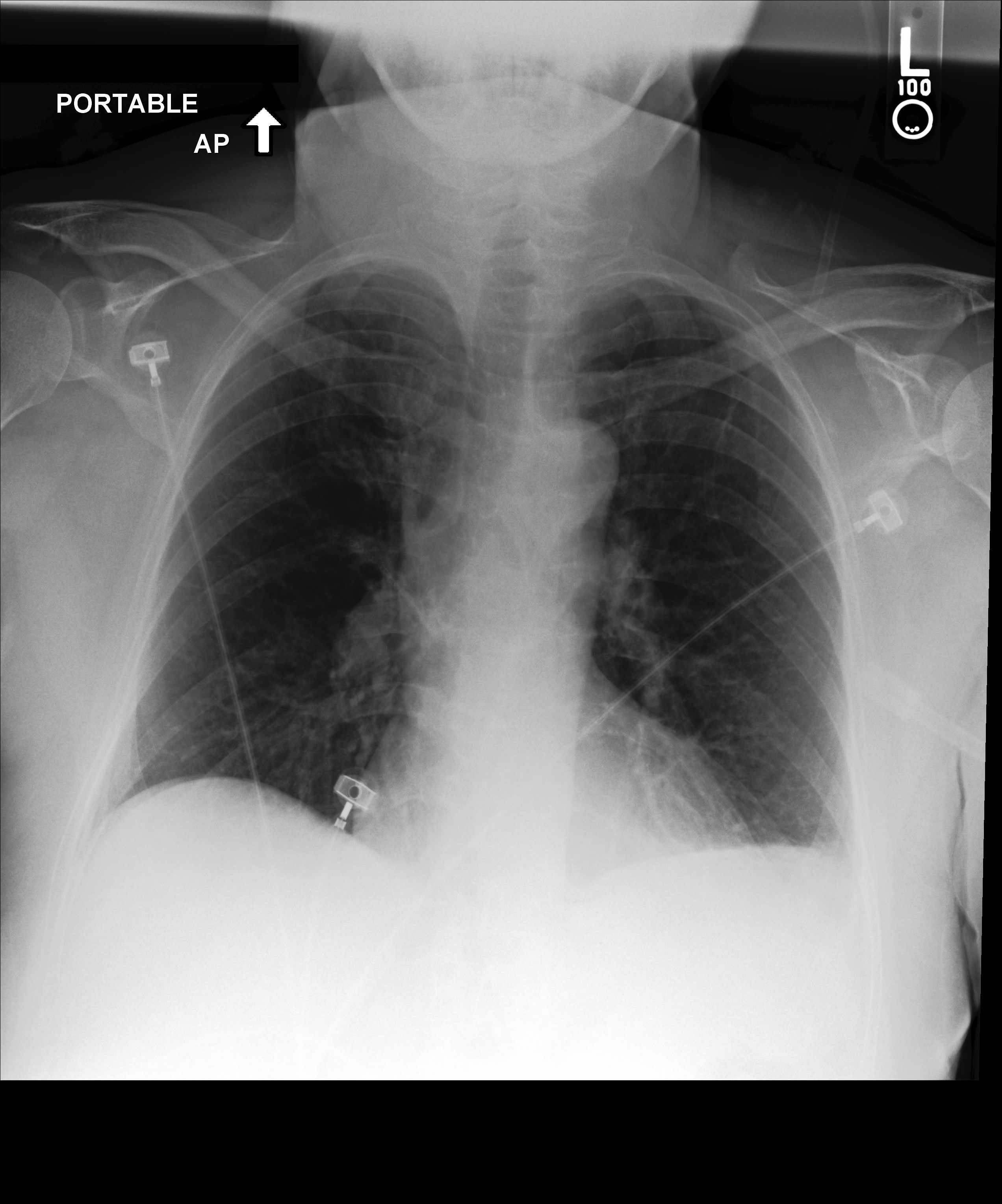}{2.25cm}{2.8cm}\end{minipage}\hfill
\begin{minipage}[t]{\dimexpr\linewidth-2.65cm\relax}\vspace{0pt}
{\scriptsize\textbf{Report.}\  \gapmark{} No airspace opacification. No pneumothorax. No pulmonary edema. Mild density seen in the left costophrenic angle which may represent atelectasis or a small pleural effusion. Suture material projecting over the right hilar area. Narrowing of the subglottic trachea is probably due to recent intubation. The cardiomediastinal shadow is unchanged. No airspace opacification. No pneumothorax. No pulmonary edema. Mild density seen in the left costophrenic angle which may represent atelectasis or a small pleural effusion. Suture material projecting over the right hilar area. Narrowing of the subglottic trachea is probably due to recent intubation.}
\par\vspace{5pt}\maskbox{The cardiomediastinal shadow is unchanged.}
{\scriptsize\setlength{\tabcolsep}{4pt}\renewcommand{\arraystretch}{1.2}
\begin{tabular}{@{}r c c@{}}
& bidirectional & left-only \\[2pt]
\textbf{diffusion} & \fillbox{hueB!12}{hueB!60}{\qualok}{The cardiomediastinal shadow is normal.} & \fillbox{hueE!8}{hueE!45}{\qualbad}{Impression: 1. No acute pulmonary process} \\[3pt]
\textbf{AR} & \fillbox{hueE!8}{hueE!45}{\qualbad}{Lungs are clear.} & \fillbox{hueE!8}{hueE!45}{\qualbad}{Bones and soft tissues are unremarkable.} \\
\end{tabular}}
\end{minipage}
}

\qframe{%
\begin{minipage}[t]{2.35cm}\vspace{0pt}\centering\qimg{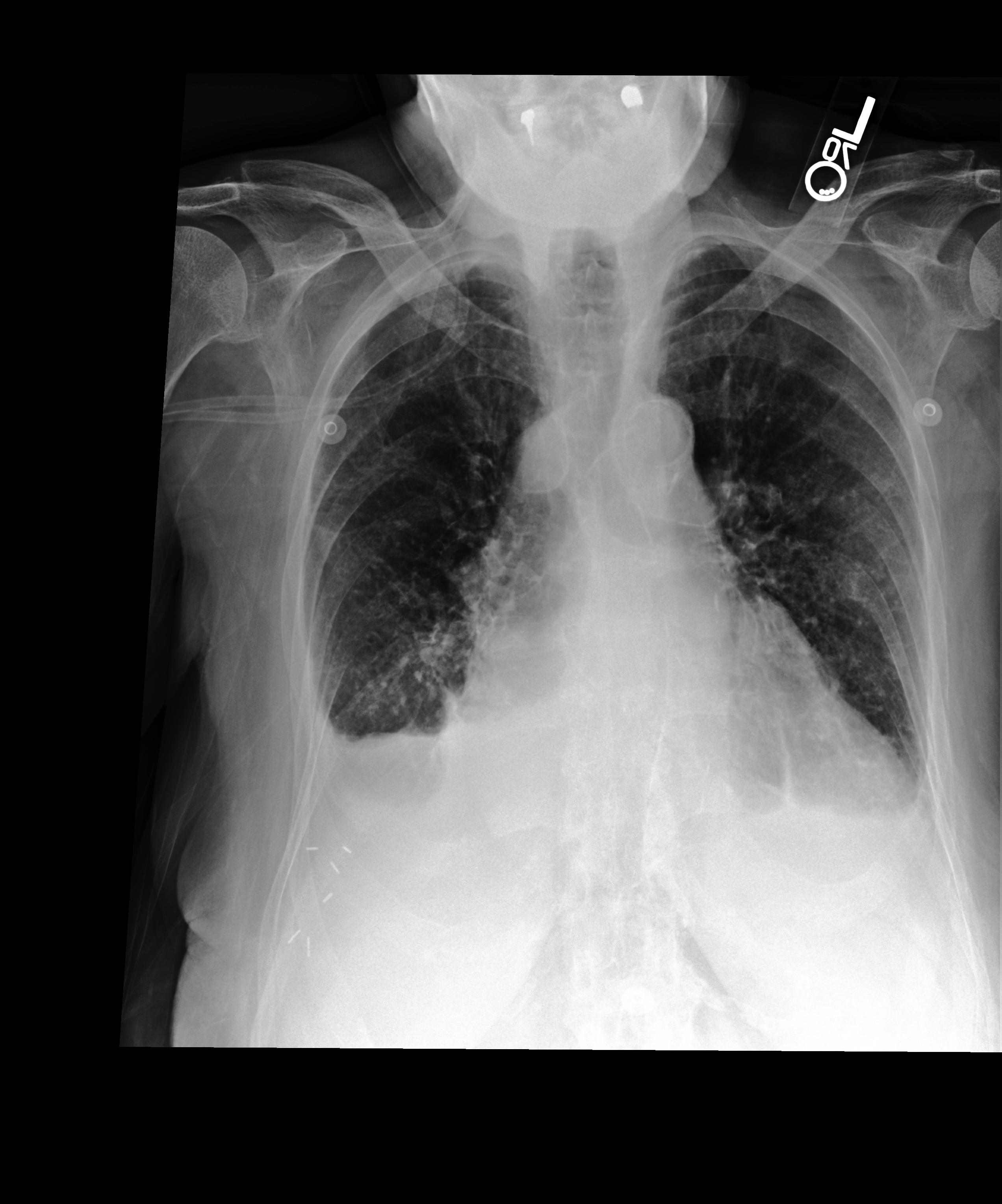}{2.25cm}{2.8cm}\end{minipage}\hfill
\begin{minipage}[t]{\dimexpr\linewidth-2.65cm\relax}\vspace{0pt}
{\scriptsize\textbf{Report.}\ There are moderate bilateral pleural effusions with overlying atelectasis, underlying basilar consolidation is not excluded in the appropriate clinical setting. The cardiomediastinal silhouette is grossly stable. \gapmark{} No evidence of pneumothorax is seen. Bilateral pleural effusions with overlying atelectasis, underlying consolidation not excluded. Possible mild vascular congestion.}
\par\vspace{5pt}\maskbox{There is minimal pulmonary vascular congestion.}
{\scriptsize\setlength{\tabcolsep}{4pt}\renewcommand{\arraystretch}{1.2}
\begin{tabular}{@{}r c c@{}}
& bidirectional & left-only \\[2pt]
\textbf{diffusion} & \fillbox{hueB!12}{hueB!60}{\qualok}{There is mild pulmonary vascular congestion.} & \fillbox{hueE!8}{hueE!45}{\qualbad}{There are surgical clips in the right} \\[3pt]
\textbf{AR} & \fillbox{hueE!8}{hueE!45}{\qualbad}{Lungs are clear without focal consolidation.} & \fillbox{hueE!8}{hueE!45}{\qualbad}{The lungs are otherwise clear without focal consolidation, suspicious pulmonary nodules, or pneumothorax.} \\
\end{tabular}}
\end{minipage}
}

\qframe{%
\begin{minipage}[t]{2.35cm}\vspace{0pt}\centering\qimg{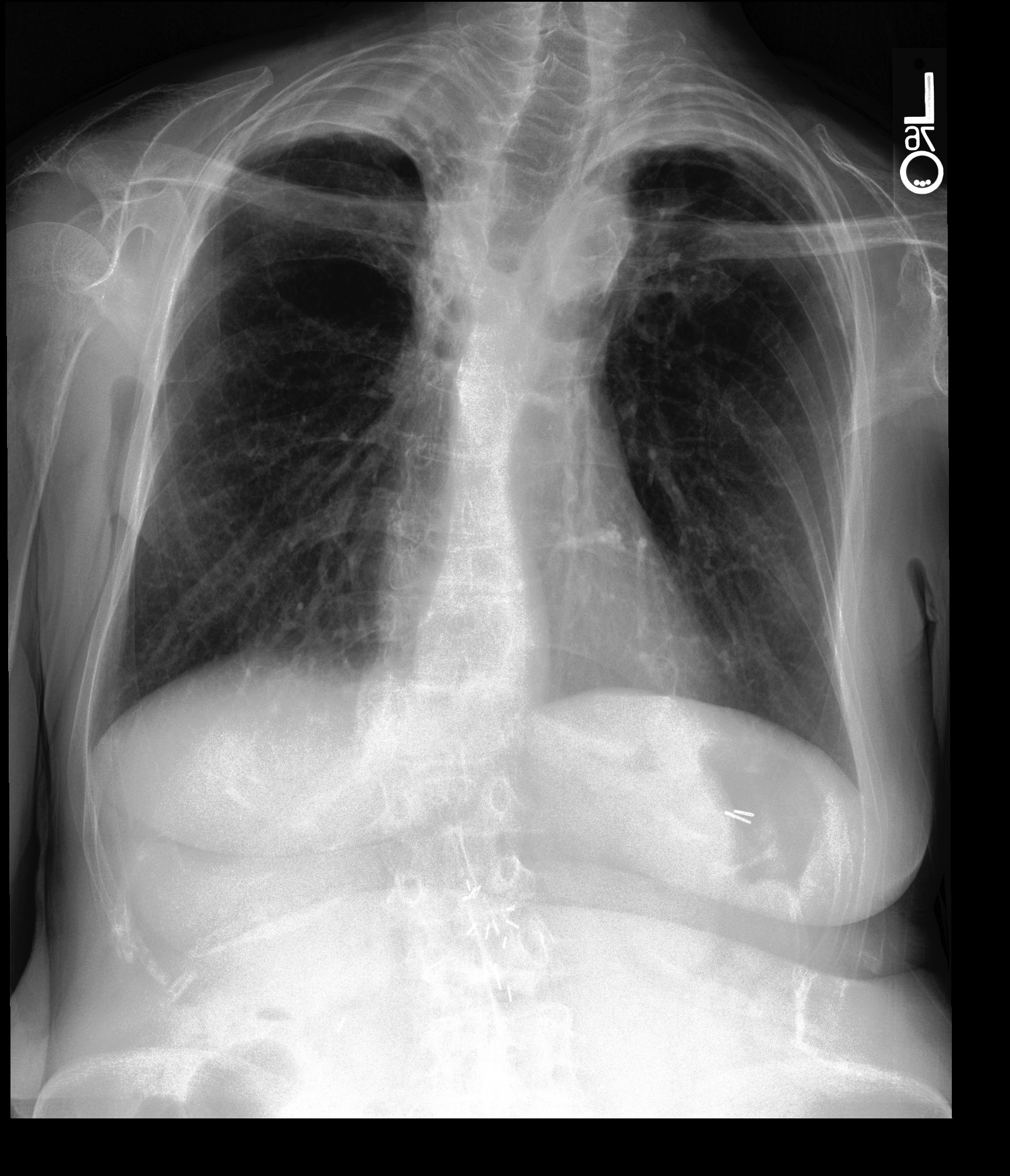}{2.25cm}{2.8cm}\end{minipage}\hfill
\begin{minipage}[t]{\dimexpr\linewidth-2.65cm\relax}\vspace{0pt}
{\scriptsize\textbf{Report.}\ The previous bilateral pleural effusions have resolved. Substantial apical thickening bilaterally and lung scarring, the sequela of likely radiation therapy is unchanged. \gapmark{} No evidence of pneumonia. Resolution of pleural effusions. Sequela of radiation induced changes including biapical scarring and fibrosis.}
\par\vspace{5pt}\maskbox{The cardiac size is normal.}
{\scriptsize\setlength{\tabcolsep}{4pt}\renewcommand{\arraystretch}{1.2}
\begin{tabular}{@{}r c c@{}}
& bidirectional & left-only \\[2pt]
\textbf{diffusion} & \fillbox{hueB!12}{hueB!60}{\qualok}{The heart size is normal.} & \fillbox{hueE!8}{hueE!45}{\qualbad}{There is no acute pulmonary consolidation} \\[3pt]
\textbf{AR} & \fillbox{hueE!8}{hueE!45}{\qualbad}{There is no new airspace opacity.} & \fillbox{hueE!8}{hueE!45}{\qualbad}{There is no new focal consolidation, pneumothorax, or pleural effusion.} \\
\end{tabular}}
\end{minipage}
}

\end{document}